\documentclass[twoside,11pt]{article}

\usepackage{blindtext}

\usepackage{xcolor}
\definecolor{citeblue}{HTML}{04146E}
\usepackage{enumitem}
\usepackage{array}
\usepackage{tabularx}
\newcolumntype{C}[1]{>{\centering\arraybackslash}p{#1}}
\newcolumntype{Y}{>{\centering\arraybackslash}X}
\usepackage{jmlr2e}

\hypersetup{
  colorlinks=true,
  citecolor=citeblue,
  linkcolor=black,
  urlcolor=citeblue
}

\usepackage{url}
\usepackage[utf8]{inputenc} 
\usepackage[T1]{fontenc}    
\usepackage{booktabs}       
\usepackage{amsfonts}       
\usepackage{nicefrac}       
\usepackage{microtype}      
\usepackage{xcolor}         
\usepackage{amsmath}
\usepackage{graphicx} 
\usepackage{mathtools}  
\usepackage{multirow}
\usepackage{makecell}
\usepackage{wrapfig}

\usepackage{caption}
\usepackage{lastpage}
\jmlrheading{xx}{2026}{1-\pageref{LastPage}}{x/xx; Revised x/xx}{x/xx}{21-0000}{Jusheng Zhang, Yijia Fan, Kaitong Cai, Keze Wang, and Wenhao Wang}

\ShortHeadings{Kolmogorov-Arnold Fourier Networks}{Kolmogorov-Arnold Fourier Networks}
\firstpageno{1}

\begin{document}

\title{Kolmogorov-Arnold Fourier Networks}

\author{
\name \hspace{-1.5mm}Jusheng Zhang \email zhangjusheng19981128@gmail.com \\
\addr Sun Yat-sen University
\AND
\name Yijia Fan \email fanyj28@mail2.sysu.edu.cn \\
\addr Sun Yat-sen University
\AND
\name Kaitong Cai \email caikt3@mail2.sysu.edu.cn \\
\addr Sun Yat-sen University
\AND
\name Keze Wang\thanks{Corresponding authors.} \email wangkz@mail.sysu.edu.cn \\
\addr Sun Yat-sen University
\AND
\name Wenhao Wang\footnotemark[1] \email wangwenhao@vastilab.com \\
\addr Vast Intelligence Lab
}

\editor{-}

\maketitle

\begin{abstract}
Although Kolmogorov-Arnold-based interpretable networks (KANs) possess strong theoretical expressiveness, they suffer from severe parameter explosion and limited ability to capture high-frequency features in high-dimensional tasks. To address these issues, we propose the Kolmogorov-Arnold Fourier Network \textbf{(KAF)}, which fundamentally redefines the KAN paradigm through spectral reparameterization. Our key contributions include:  \textbf{(1)} proposing a fundamental basis transformation from the local, grid-based B-spline representation to a global, adaptive spectral representation. This shift changes the network's inductive bias, reducing parameter complexity from $O(G)$ to $O(1)$ while preserving expressiveness; \textbf{(2)} introducing trainable Random Fourier Features (RFF) initialized via a spectral alignment strategy, which allows the model to break the smoothness limitation of fixed kernels and accurately capture high-frequency components; and \textbf{(3)} implementing an adaptive hybrid GELU-Fourier activation mechanism that progressively enhances frequency representation during training. Comprehensive experiments demonstrate the superiority of KAF across computer vision (CV), natural language processing (NLP), audio, and partial differential equation (PDE) solving tasks, achieving state-of-the-art performance with improved efficiency. The code is available at \url{https://github.com/kolmogorovArnoldFourierNetwork/KAF}.

\end{abstract}


\section{Introduction}
The interpretability of deep neural networks \citep{sjwl01,sjwl02} has long been one of the core challenges in the field of machine learning. The Kolmogorov-Arnold \citep{kan,KA1933} theorem states that any continuous multivariate function can be represented through a combination of univariate functions \citep{GW,GWBJ}. This theory provides significant inspiration for the design of interpretable neural network architectures. Based on this theory, Kolmogorov-Arnold Networks (KAN) \citep{kan,zheKAN} have been proposed, which replace the fixed activation functions in traditional multilayer perceptrons (MLPs \citep{MLP}) with learnable B-spline \citep{B-spline} basis functions, theoretically demonstrating strong expressive potential and flexibility. By introducing trainable nonlinear activation functions, KAN enables the network to dynamically adjust the shape of the activation functions according to the characteristics of the data, thereby enhancing the adaptability and performance of the model.

However, despite the significant theoretical advantages of KAN, its practical application faces two fundamental issues that severely limit its generalization and adoption in high-dimensional tasks.
\textbf{(1) Inefficient parameter utilization.}
The dual-matrix architecture of KAN, namely the activation function matrix and the B-spline coefficient matrix, leads to a rapid increase in the number of parameters. Compared to traditional MLPs, where the parameter count scales with input $\times$ output plus bias, KAN's parameter count grows several times larger. This makes it challenging to apply KAN to high-dimensional tasks such as computer vision. The explosion in parameters \citep{gaoweipingyu,GW,GWBJ} not only increases storage and computational costs, but also significantly reduces the efficiency of both training and inference \citep{canshu,canshu2}.
\textbf{(2) Limited spectral representation.}
The B-spline basis functions employed by KAN \citep{B-spline,byt,byt3} exhibit inherent spectral limitations when performing function approximation in high-dimensional spaces. Their smoothness makes it difficult to accurately capture high-frequency components, leading to suboptimal performance on data with rich spectral features, such as natural images or audio waveforms. This limitation adversely affects model performance and stability in practical applications \citep{gaoweipingyu}.
\textbf{Together}, these limitations create a paradox between theory and practice: although KAN theoretically encompasses the functionality of MLPs, its inefficiency and spectral distortion force practitioners to trade interpretability for scalability.

\begin{wrapfigure}{r}{0.6\textwidth}
  \centering
  \vspace{-6mm}
  \includegraphics[width=0.6\textwidth]{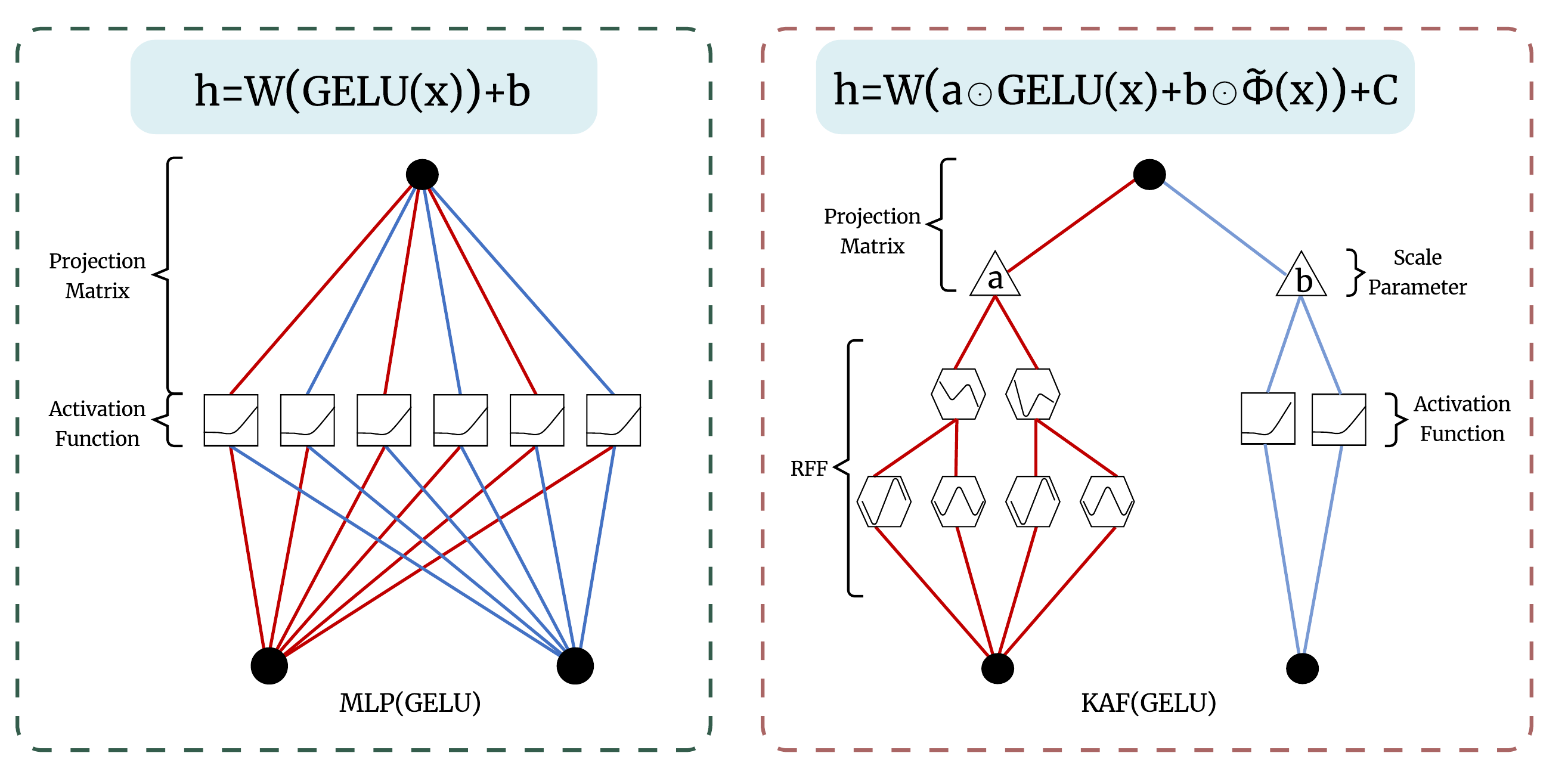}
  \vspace{-9mm}
\caption{Comparison between a standard GELU-MLP layer and the proposed GELU-KAF layer. KAF augments the GELU branch with trainable Random Fourier Features (RFF) and learnable scaling parameters, enabling more flexible spectral feature transformations.}
\vspace{-4mm}
   \label{fig:comppipeline}
\end{wrapfigure}

To address the aforementioned issues, the key challenge is balancing the inherent trade-off between model interpretability and parameter efficiency, which has long plagued traditional neural networks. This paper makes an attempt to fundamentally redefine the traditional KAN paradigm~\citep{fuliyekaishan}.  Specifically, this paper introduces an innovative neural network architecture, \textbf{Kolmogorov-Arnold-Fourier Networks (KAF)}, which employs Fourier domain reparameterization and dynamic activation evolution, aiming to bridge the gap between interpretability and parameter efficiency. 
Fig.~\ref{fig:comppipeline} illustrates the difference between a standard GELU-MLP layer and the proposed KAF layer.

Our main contributions include:
\begin{itemize}[leftmargin=1.3em,itemsep=1pt,topsep=2pt]
    \item By leveraging the associative property of matrix multiplication, we merge the two large matrices ($WA$ and $WB$) of KAN, thereby reducing the parameter complexity from $O(d_{\text{in}} \times d_{\text{out}} \times (G + K + 3))$ to $O(d_{\text{in}} \times d_{\text{out}})$, while preserving the expressive power of the model. This approach not only effectively reduces the number of parameters but also enhances the model's scalability in high-dimensional tasks.
    \item We replace the traditional B-spline basis functions with trainable Random Fourier Features (RFF) to eliminate the need for the spline coefficient matrix. An initialization strategy based on the Central Limit Theorem ($\sigma = 1.64$) aligns the RFF spectrum with the prior knowledge of natural signals, avoiding spectral leakage issues. This significantly enhances the model's spectral fidelity and expressive power in high-dimensional spaces.
    \item We design a hybrid GELU-Fourier activation function with learnable coefficients $\{a, b\}$. During training, these coefficients are dynamically adjusted through gradient backpropagation, enabling an automatic transition from fixed activation functions to hybrid Fourier-symbolic representations. This dynamic activation mechanism not only optimizes the model's learning trajectory but also ensures the stability and efficiency of the final Fourier-driven inference mode.
\end{itemize}

\section{Related Work}
\textbf{Multi-Layer Perceptrons and Current Challenges.} The design and optimization of deep learning \citep{deit} models remain central to machine learning research. Traditional MLPs \citep{MLP}, among the earliest neural networks \citep{sjwl02}, offer simplicity and scalability, with rich theoretical foundations. While ResNet \citep{resnet18} and Transformer \citep{transform} models have shown remarkable performance across various tasks, MLPs face challenges in theoretical interpretability and practical bottlenecks. Traditional activation functions like ReLU \citep{ReLU1,ReLU2} and Sigmoid \citep{Sigmoidjihuo} often fail to adapt to complex data, and despite their efficiency, MLPs struggle with high-frequency features and complex distributions. Improving activation mechanisms and parameter efficiency has become crucial for enhancing MLPs' adaptability to high-dimensional data.

\textbf{Kolmogorov-Arnold Networks and Scalability Issues.} The Kolmogorov-Arnold \citep{2019duichengkanl} Theorem underpins networks for approximating continuous multivariable functions. The pioneering KAN replaced fixed activations with B-spline \citep{B-spline} functions but faces challenges in high-dimensional applications due to parameter explosion and GPU inefficiency. Recent improvements include KAN-Transformer, MLP-KAN with sparse parameters, and FAN \citep{FAN} with Fourier activations, all seeking to balance interpretability with scalability.

\textbf{Enhancing Spectral Representation with KAF.} To address high-frequency modeling challenges, Random Fourier Features (RFF \citep{suijifuly}) enable spectral domain mapping, with variants like Learnable RFF and SIREN enhancing expressiveness. Our proposed KAF incorporates GELU and learnable Fourier features, with scale factor control and variance initialization. This reduces parameters while improving spectral representation. KAF maintains KAN's interpretability while enhancing scalability and efficiency, showing superior performance in capturing high-frequency \citep{SIRENgaoping} details across NLP, vision, audio, and traditional machine learning tasks.

\section{Methodology}

This section presents the proposed Kolmogorov-Arnold Fourier Network (KAF). We first review the Kolmogorov-Arnold theorem and the standard KAN formulation to identify the sources of parameter growth and spectral limitations, and then introduce KAF, which replaces spline-based edge functions with trainable Random Fourier Features and a GELU-Fourier hybrid activation.

\subsection{Kolmogorov-Arnold Theorem}

The Kolmogorov-Arnold \citep{fuliyekaishan,KA1933} theorem, proposed by Soviet mathematicians Vladimir Arnold and Andrey Kolmogorov in the 1950s, states that any continuous \citep{2020jinsi} multivariate function \( f: [0,1]^d \rightarrow \mathbb{R} \) can be represented as a superposition of univariate functions:
\begin{equation}
f(x_1, x_2, \dots, x_d) = \sum_{q=1}^{2d+1} \Phi_q \left( \sum_{p=1}^{d} \phi_{q,p}(x_p) \right),
\end{equation}
where \(\Phi_q: \mathbb{R} \rightarrow \mathbb{R}\) and \(\phi_{q,p}: [0,1] \rightarrow \mathbb{R}\) are univariate continuous functions. This theorem provides a theoretical foundation for dimensionality reduction in high-dimensional function approximation.

In \citep{shuxue1}, the theorem suggests that high-dimensional functions can be captured through low-dimensional \citep{gaoweipingyu} transformations, resembling the hierarchical structure of neural networks. However, compared to traditional neural networks, the number of parameters required by the Kolmogorov-Arnold theorem may lead to a more lengthy and resource-consuming training process. Due to the non-smoothness of certain low-dimensional functions and the difficulties in training optimization, this theorem has not been practically applied in the field of neural networks for a long time.

\subsection{Kolmogorov-Arnold Network (KAN)}

Although the Kolmogorov-Arnold \citep{fuliyekaishan} theorem was proposed quite early, (KAN \citep{kan}) is proposed according to this theorem, demonstrating that this structure can, in a sense, serve as an alternative to traditional MLP models. In the KAN network, each layer can be represented by the following formula:
\begin{equation}
f(\mathbf{x}) = \Phi \circ \mathbf{x} = \left[ \sum_{i=1}^{d_{in}} \phi_{1,i}(x_i) \quad \cdots \quad \sum_{i=1}^{d_{in}} \phi_{d_{out},i}(x_i) \right],
\end{equation}
where \(\Phi\) is a matrix of basis functions. This formula aligns with the form of the Kolmogorov-Arnold theorem. However, in practical applications, they chose B-spline \citep{yangtiao} basis functions as the basis functions $\phi_{q,p}$, and added an external activation function $SILU$ to guide the update of the KAN layer \citep{rsjihuo,Sigmoidjihuo}. The formula can be expressed as
\begin{equation}
\phi(x) = w_h \operatorname{silu}(x) + w_s \operatorname{spline}(x),
\quad
\text{where }
\operatorname{spline}(x) = \sum_i c_i B_i(x).
\end{equation}
Among them, $\Phi$ represents the basis function matrix, where B-spline basis functions and the SiLU activation function were used. However, KAN suffers from excessive parameter growth, with a parameter count of $d_{\text{out}} \times d_{\text{out}} \times (G + K + 3) + d_{\text{out}}$, far exceeding MLP’s $d_{\text{in}} \times d_{\text{out}} + d_{\text{out}}$, while also being computationally inefficient on GPUs and failing to capture high-frequency components, limiting its practical applicability.

\subsection{Kolmogorov-Arnold Fourier Network (KAF)}
In the previous discussion, we pointed out that traditional networks based on the Kolmogorov-Arnold theorem (KAN) often face multiple challenges in practical applications. To address these issues, we propose an alternative approach, \textit{i.e.},  Kolmogorov-Arnold Fourier Network (KAF). By replacing B-spline basis functions with Random Fourier Features (RFF \citep{fuliye2,fuliyebiaozhun,fuliyekaishan}), which are more efficient for GPU acceleration, and introducing hybrid spectral correction for the activation functions, the network retains the advantages of the Kolmogorov-Arnold theory while achieving training efficiency and inference speed closer to that of MLPs \citep{MLP}. This section provides a detailed explanation of the overall architecture of the KAF network, the utilization of Random Fourier Features within the network, the design and scaling principles of the GELU-Fourier hybrid activation function, as well as the RFF weight initialization strategy and the theoretical justification for $\sigma=1.64$.

\textbf{Overall Architecture.}
In the overall framework of KAF, we follow the core idea of the Kolmogorov-Arnold theorem, which approximates high-dimensional target functions through the composition of several low-dimensional learnable functions. Unlike the KAN network, which directly utilizes B-spline basis functions, KAF employs Random Fourier Features (RFF) in each layer to perform a nonlinear mapping of the input, and then uses linear transformations to achieve the composition of the "outer function" and the "inner function." Specifically, the core computational process of each KAF layer can be formulated as:
\begin{equation}
\mathbf{h}^{(l)} = 
\underbrace{\mathbf{W}^{(l)}}_{\text{outer function}}
\left(
\underbrace{\mathbf{a}^{(l)} \odot \text{GELU}(\tilde{\mathbf{x}}^{(l)}) 
+ \mathbf{b}^{(l)} \odot \tilde{\phi}(\tilde{\mathbf{x}}^{(l)})}_{\text{inner function composition}}
\right) 
+ \mathbf{c}^{(l)},
\end{equation}
where:  
$\tilde{\mathbf{x}}^{(l)} = \text{LayerNorm}(\mathbf{x}^{(l)})$ is the normalized input at layer $l$;  
$\tilde{\phi}(\cdot)$ represents the nonlinear mapping based on Random Fourier Features (RFF) (detailed in Section 3.3.2);  
$\mathbf{a}^{(l)}, \mathbf{b}^{(l)} \in \mathbb{R}^n$ are learnable scaling parameters, used to modulate the contributions of GELU activation and RFF features, respectively;  
$\mathbf{W}^{(l)} \in \mathbb{R}^{m \times n}$ is the linear transformation weight, and $\mathbf{c}^{(l)} \in \mathbb{R}^m$ is the bias term.  

By stacking multiple layers of the above transformation, the KAF network constructs an efficient multi-layer approximation structure. Since RFF has excellent parallelism on GPUs, this structure avoids the high computational burden of B-spline basis functions, significantly improving training and inference efficiency while maintaining strong function approximation capabilities.

\textbf{Random Fourier Features (RFF).}
Given an input space $\mathcal{X} \subseteq \mathbb{R}^d$, we define the Random Fourier Feature (RFF \citep{fuliyebiaozhun}) mapping as a learnable embedding from the input space to a Reproducing Kernel Hilbert Space (RKHS \citep{RKHS,RKHS2,RKHS3}). For any input vector $x \in \mathcal{X}$, the feature mapping is formally defined as:
\begin{equation}
z(x;W,b)=\sqrt{\frac{1}{m}}\big[\cos(\langle x,W\rangle+b)\oplus\sin(\langle x,W\rangle+b)\big]\in\mathbb{R}^{2m},
\end{equation}
where $W\in\mathbb{R}^{d\times m}$ and $b\in\mathbb{R}^m$.
Here, $\langle \cdot, \cdot \rangle$ denotes the Euclidean inner product, and $\oplus$ represents the vector concatenation operation. The frequency matrix $W = [w_1, \dots, w_m]$ is initialized according to an input-dimension-adaptive spectral distribution: $w_{ij} \sim \mathcal{N}(0, \sigma^2/d)$, where $\sigma^2$ represents the empirical variance of the input data. The phase shift $b$ is sampled from a uniform distribution $b_i \sim \mathcal{U}[0,2\pi]$, which ensures phase diversity, a crucial property for capturing local features of signals. See Appendix~\ref{Appendix B} for more information on RFF convergence, gradient calculation, and initialization strategies.

This mapping comes with the following theoretical guarantees:

\begin{itemize}[leftmargin=1.3em,itemsep=1pt,topsep=2pt]

\item Translation Invariance: For any $x, y \in \mathcal{X}$, as $m \to \infty$, we have $\mathbb{E}[z(x)^T z(y)] \to e^{-\frac{\|x-y\|^2}{2\sigma^2}}$.

\item Differentiability: The partial derivatives $\frac{\partial z}{\partial W}$ and $\frac{\partial z}{\partial b}$ have analytical expressions, enabling end-to-end differentiation.
\end{itemize}

\textbf{GELU-Fourier Hybrid Activation.}
\textit{Design Motivation}: To balance low-frequency smoothness and high-frequency representation capability, we propose a hybrid activation function:
\begin{equation}
\mathcal{H}(\boldsymbol{x}) = 
\underbrace{\alpha \odot \mathrm{GELU}(\boldsymbol{x})}_{\mathclap{\text{Low-Frequency Basis}}} 
\mathrel{\hphantom{=}} + \mathrel{\hphantom{=}}
\underbrace{\beta \odot \boldsymbol{V}\psi(\boldsymbol{x})}_{\mathclap{\text{High-Frequency Correction}}}
\end{equation}
where $\alpha, \beta \in \mathbb{R}^d$ are learnable channel-wise scaling factors, $\boldsymbol{V} \in \mathbb{R}^{d \times 2k}$ is the frequency-domain projection matrix, and $\odot$ represents element-wise multiplication.

\textbf{Initialization Strategy of KAF.}
\begin{equation}
\scalebox{1}{$
\alpha^{(0)} \gets \boldsymbol{1}, \quad 
\beta^{(0)} \gets \epsilon \boldsymbol{1}, \quad (\epsilon = 10^{-2}), \quad
\boldsymbol{V}_{ij}^{(0)} \sim \mathcal{N}(0, 0.01)
$}
\end{equation}
The dynamic property of this initialization manifests in the following way: At the early stage of training, the small initialization of the high-frequency component $\beta$ ensures that its norm is much smaller than that of the low-frequency component $\alpha$, prioritizing the learning of low-frequency features. As training progresses, the natural growth of weights allows the norm of $\beta$ to increase approximately proportionally to the training time $t$, thereby gradually enhancing the representation of high-frequency features.

\textbf{Implementation of the Kolmogorov-Arnold Architecture and RFF Initialization.}

\textit{Theorem Definition}: The Kolmogorov-Arnold representation theorem states that any continuous function $f \in C([0,1]^d)$ can be expressed as a finite composition of univariate functions:
\begin{equation}
f(\boldsymbol{x}) = \sum_{q=0}^{2d} \Phi_q\left( \sum_{p=1}^d \phi_{q,p}(x_p) \right),
\end{equation}
where $\phi_{q,p}: \mathbb{R} \to \mathbb{R}$ are univariate nonlinear functions, and $\Phi_q: \mathbb{R} \to \mathbb{R}$ are composition functions.

\textit{Architecture Implementation}: We modularize the neural network to efficiently approximate this mapping, establishing the following correspondences:
\begin{equation}
\begin{aligned}
\phi_{q,p}(x_p) 
    &\mapsto 
    \underbrace{\mathrm{GELU}\bigl(w_p^{(q)}x_p + b_p^{(q)}\bigr)}_{\mathclap{\text{Low-Frequency Basis}}} 
    + \boldsymbol{\beta}_q^\top \mkern-2mu 
    \underbrace{\psi_{\text{RFF}}(x_p)}_{\mathclap{\text{High-Frequency Basis}}}, 
    \\[1ex]  
\Phi_q(\cdot) 
    &\mapsto 
    \boldsymbol{\alpha}_q^\top \mkern-2mu \text{Linear}(\cdot).
\end{aligned}
\end{equation}
Here, $\psi_{\text{RFF}}(x_p) = [\cos(\omega_1x_p+\theta_1), \sin(\omega_1x_p+\theta_1), \dots]$ represents the Random Fourier Features (RFF), and $\boldsymbol{\alpha}_q, \boldsymbol{\beta}_q \in \mathbb{R}^k$ are learnable modulation parameters.

\textit{Spectral Complementarity Mechanism}:
(1) GELU Properties: The activation function $\sigma(wx+b)$ provides a smooth gating effect in the low-frequency domain, satisfying $\mathbb{E}[\sigma(wx)] \propto \mathcal{N}(0,1/\sqrt{2})$.
(2) RFF Enhancement: The use of mixed-frequency bases $\{\cos(\omega_mx+\theta_m)\}_{m=1}^M$ expands spectral coverage.
(3) Dynamic Balancing: The learnable parameters $\boldsymbol{\alpha}, \boldsymbol{\beta}$ enable an adaptive trade-off:
\begin{equation}
\tilde{\phi}(x) = \alpha \cdot \mathrm{GELU}(x) + \beta \cdot \psi_{\text{RFF}}(x),
\end{equation}
where the initial values are set to $\alpha^{(0)}=1$ and $\beta^{(0)}=10^{-2}$ to ensure training stability.

\textit{RFF Initialization Strategy}: To fully leverage spectral complementarity, we adopt a refined initialization scheme:
\begin{itemize}[leftmargin=1.3em,itemsep=1pt,topsep=2pt]
    \item Frequency Matrix $\boldsymbol{W}$: To ensure spectral balance and avoid bias towards low or high frequencies, we initialize $\boldsymbol{W}$ using a scaled normal distribution \citep{chushihua,chushihua2}:
\begin{equation}
\omega_{ij} \sim \mathcal{N}\left(0,\, \frac{\gamma}{\sqrt{d_{\text{in}} \cdot \mathbb{E}[\|\sigma(x)\|^2]}}\right),
\end{equation}
This initialization is designed to align with the spectral distribution of the input data. The denominator normalizes the standard deviation based on input dimensionality $d_{\text{in}}$ and the expected squared norm of the activation function $\mathbb{E}[\|\sigma(x)\|^2]$, ensuring a stable variance propagation during training. For the GELU activation function, why $\sigma(x)=1.64$ will be proved later in Appendix~\ref{prof:1.64}.

\item  Phase Shift $\boldsymbol{b}$: Uniformly sampled to cover a complete period,
\begin{equation}
b_i \sim \mathcal{U}(0,\,2\pi).
\end{equation}
\item  Linear Projection Layer: Initialized using Xavier initialization,
\begin{equation}
\boldsymbol{V}_{ij} \sim \mathcal{U}\left(-\sqrt{6/(d_{\text{in}}+d_{\text{out}})},\,\sqrt{6/(d_{\text{in}}+d_{\text{out}})}\right).
\end{equation}
\end{itemize}

\begin{table}[!t]
    \centering
    \caption{Comparison of parameter count and FLOPs per layer for KAN, KAF, and MLP models.}
    \vspace{-1mm}
    \label{tab:param_flops}
    \small
    \begin{tabular}{l|l|l}
        \hline
        \textbf{Model} & \textbf{Param Count (Single Layer)} & \textbf{FLOPs (Single Layer)} \\
        \hline
        \textbf{KAN} 
        & 
        $ d_{\mathrm{in}} d_{\mathrm{out}} (G+K+3) + d_{\mathrm{out}} $
        & 
        $7d_{\mathrm{in}} + (d_{\mathrm{in}} d_{\mathrm{out}})\left[9K (G + 1.5K) + 2G - 2.5K + 3\right]$ \\
        \textbf{KAF} 
        & 
        $ d_{\mathrm{in}} M + M + 2d_{\mathrm{in}} + d_{\mathrm{in}}d_{\mathrm{out}} + d_{\mathrm{out}} $
        & 
        $ 4d_{\mathrm{in}}M + 2d_{\mathrm{in}} + 2d_{\mathrm{in}}d_{\mathrm{out}} + 5d_{\mathrm{in}} $ \\
        \textbf{MLP} 
        & 
        $ d_{\mathrm{in}} d_{\mathrm{out}} + d_{\mathrm{out}} $
        & 
        $ 2 d_{\mathrm{in}} d_{\mathrm{out}} + 5 d_{\mathrm{out}} $ \\
        \hline
    \end{tabular}
    \vspace{-3mm}
\end{table}

\textbf{Parameter and FLOPs Comparison.}
To evaluate the scale of parameters and computational overhead of KAF, we compare the number of parameters and floating-point operations (FLOPs) for KAF, KAN, and MLP in a single layer setting.

Table~\ref{tab:param_flops} summarizes the parameter count and FLOPs for each model. KAN exhibits the highest parameter count due to its recursive B-spline computations, while KAF, by leveraging Random Fourier Features (RFF), achieves a balance between parameter efficiency and spectral representation. MLP remains the simplest in terms of computation.
For the detailed derivation of these calculations, please refer to Appendix~\ref{Appendix A}.

\section{Experiments}

The objective of this section is to evaluate the performance of mainstream models when their MLP \citep{MLP} or KAN components are replaced with KAF. We conduct experiments across a variety of tasks, including visual classification, natural language processing, audio classification, traditional machine learning benchmarks, function fitting, and differential equation solving. These evaluations use representative architectures such as ResNet-18 \citep{resnet18}, DeiT \citep{deit}, MLP-Mixer \citep{MLP-Mixer}, and GPT-2 \citep{GTP2}. Unless otherwise specified, all experiments use the Adam optimizer \citep{kingmaadam}, with learning rates selected according to the task. All experiments are conducted on an NVIDIA RTX 4090D GPU.

\subsection{Comprehensive Evaluation Based on Kanbefair}
Based on Kanbefair \citep{kanbase}, we conduct a comprehensive evaluation of KAF on vision \citep{VIT}, NLP \citep{GTP2}, audio, and machine learning tasks to compare its performance with existing models. We selected MLP (with GELU activation), KAN, FAN \citep{FAN}, and GPKAN \citep{KAT} for experimentation. Note that we use the original input by default in all experiments, leaving layernorm disabled.

\textbf{Experimental setup.}
To account for differences in model convergence speeds, KAN was trained for 100 epochs, while other models were trained for 40 epochs. During training, the maximum test accuracy was recorded as the primary evaluation metric, focusing on classification tasks across 8 vision datasets, 2 NLP datasets, 2 audio datasets, and 4 additional machine learning datasets. For all models, hidden layer sizes were explored in the range of 2, 4, 8, 16, 32, 64, 128, 256, 512, and 1024. For KAF, the key parameters included 9 grids, an activation expectation of 1.64, and GELU as the activation function. For KAN, the number of B-spline grids was set to 3, 5, 10, or 20; B-spline degrees were configured as 2, 3, or 5; and the B-spline range was [-1, 1], [-2, 2], or [-4, 4]. For MLP, we experimented with both GELU \citep{gelus} and ReLU \citep{ReLU1,ReLU2} activations. FAN’s p\_ratio was set to 0.25 to regulate the proportion of preserved information during transformation, and GPKAN used GELU-based initialization for enhanced convergence.

\begin{figure*}[t]
    \centering
    \includegraphics[width=1\linewidth]{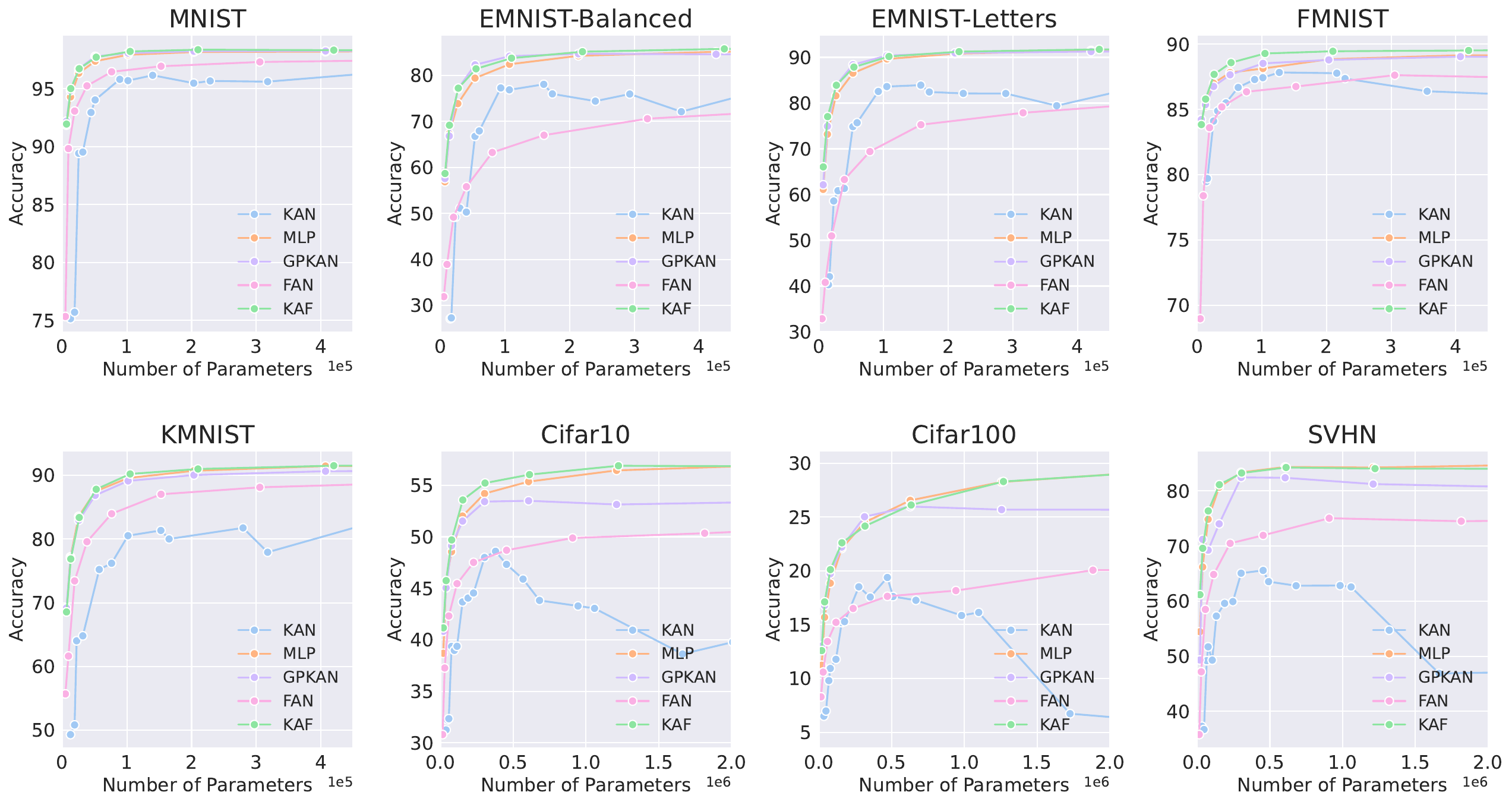}
    \vspace{-3mm}
    \caption{Comparison of different models (KAN, MLP, GPKAN, FAN, and KAF) across multiple datasets, including MNIST, EMNIST, FMNIST, KMNIST, CIFAR-10, CIFAR-100, and SVHN. The results demonstrate that KAF consistently achieves higher accuracy while using fewer parameters.}
    \vspace{-4mm}
    \label{fig:accuracy_params_vision}
\end{figure*}

\begin{figure}[t]
    \centering
    \includegraphics[width=1.0\linewidth]{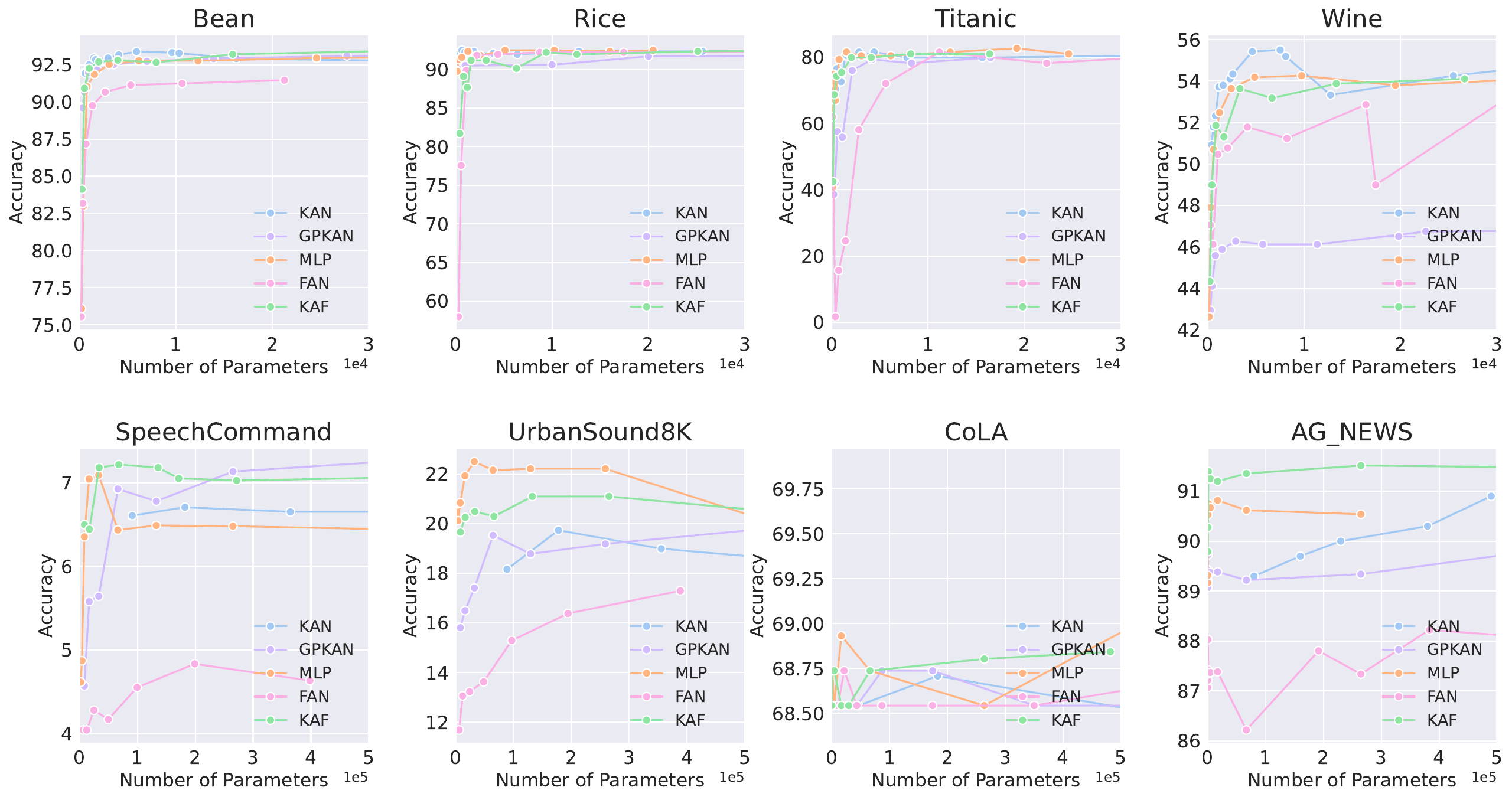}
    \vspace{-3mm}
    \caption{Comparison of KAN, GPKAN, MLP, FAN, and KAF across NLP, audio, and traditional machine learning datasets. KAF consistently achieves higher accuracy with fewer parameters, especially on datasets such as Bean, Rice, and AG News.}
     \vspace{-4mm}
    \label{fig:accuracy_params_ml}
\end{figure}

\textbf{Experimental results.}
As shown in Fig. \ref{fig:accuracy_params_vision}, we conduct a systematic comparison of KAF and several baseline models (e.g., ResNet, ViT, MLP-Mixer) on vision datasets, including MNIST \citep{lecun1998gradient}, EMNIST \citep{emnist}, KMNIST \citep{clanuwat2018deep}, CIFAR 10/100 \citep{krizhevsky2009learning}, and SVHN \citep{netzer2011reading}. 
The results in Fig. \ref{fig:accuracy_params_vision} demonstrate that KAF consistently achieves the highest accuracy under the same parameter settings across different scales. Notably, on more challenging tasks such as CIFAR-10 and SVHN, KAF exhibits significant accuracy improvements compared to other models. These results highlight KAF's robustness in addressing high-dimensional data challenges.
In addition to vision tasks, we evaluate KAF on NLP, audio, and traditional machine learning datasets. As shown in Fig.~\ref{fig:accuracy_params_ml}, KAF consistently achieves higher accuracy than the baseline models under comparable parameter budgets, especially on datasets such as Bean, Rice, and AG News. These results indicate that KAF is not only effective for visual recognition but also generalizes to text, audio, and tabular learning tasks.

\subsection{Experiments on Using KAF Components in Complex Vision Models}
To comprehensively evaluate the performance of KAF in large-scale vision models after replacing corresponding layers, we assess its impact on accuracy, computation time, and generalization across various models. In this section, we conduct comparative experiments on commonly used large-scale vision models, including ResNet-18, ViT-Tiny, and MLP-Mixer-S/16, as well as the latest model incorporating KAN components, MLP\_KAN (based on DeiT). In each case, we replace the original KAN or MLP modules with KAF, KAN, MLP, GPKAN, and FAN, respectively, to analyze their performance differences.

\textbf{Experimental setup.}
The experiments utilize CIFAR-10, CIFAR-100 \citep{krizhevsky2009learning}, and ImageNet-1K for training and testing. ResNet-18 and MLP-Mixer-S/16 are trained for 100 epochs, while ViT-Tiny \citep{Kan_mlp,VIT} and MLP\_KAN are trained for 300 epochs. All other training hyperparameters follow the official recommended settings for each model.
Both KAF and MLP use GELU as the activation function; for KAN, we set the grid size to 5 and the B-spline degree to 3; FAN’s p\_ratio is fixed at 0.25; GPKAN adopts GELU-based initialization; LayerNorm is disabled by default; and Dropout follows each model’s standard setting.

\textbf{Experimental results.}
Table~\ref{tab:model_comparison} summarizes performance across multiple datasets when various feature Mixers replace the original modules. KAF generally meets or exceeds baseline accuracy while maintaining a reasonable parameter count and computational cost. For instance, in ResNet-18 on CIFAR-10, KAF achieves 91.72\% Top-1 accuracy (compared to MLP’s 91.19\%), and on ImageNet-1K, it improves MLP-Mixer from 63.5\% to 64.7\%. Although compact approaches like FAN can reduce parameters, they often yield lower accuracy (e.g., 58.2\% on ImageNet-1K). Similarly, with ViT-T/16, KAF’s moderate parameter increment still outperforms MLP, while in MLP\_KAN (DeiT), KAF boosts baseline accuracy from 49.0\% to 53.8\%.
Overall, KAF demonstrates gains and stable training on challenging tasks and architectures, including attention-rich models. These results highlight its potential as a more efficient and effective Mixer replacement in modern vision networks, balancing parameter overhead with performance improvements.

\subsection{Experiments on LLMs with KAF Components}
To evaluate the potential of KAF in language models, we integrate it into the GPT-2 architecture by replacing the Feed-Forward Network (FFN)'s MLP with KAF or KAN. We then train and evaluate the models on large-scale text datasets, assessing their impact on language modeling quality and model complexity.

\textbf{Experimental setup.}
We conduct experiments using OpenWebText and WikiText, two widely used text datasets. The base model is GPT-2 (Small version), where the two-layer MLP in the Feed-Forward Network (FFN) is replaced with KAF or KAN while maintaining the same parameter scale.
All other Transformer configurations \citep{transform}, including multi-head attention, token embeddings, and positional encoding, remain consistent with the official GPT-2 implementation.

\begin{table}[t]
\centering
\caption{Comparison of different feature mixers in common vision architectures. Parameters refer to the total model size, and FLOPs are computed for forward propagation.}
\small
\vspace{-1mm}
\textbf{(a) ResNet-18 on CIFAR-10 and ViT-T/16 on ImageNet-1K}

\vspace{1mm}
\begin{tabularx}{\textwidth}{C{3cm} C{3cm} C{2cm} Y Y Y}
\toprule
\textbf{Model} & \textbf{Dataset} & \textbf{Mixer} & \textbf{\#Param.} & \textbf{FLOPs} & \textbf{Top-1} \\
\midrule
ResNet/18 & CIFAR-10 & MLP & 11.1M & 0.56G & 91.19 \\
ResNet/18 & CIFAR-10 & KAF & 12.0M & 0.63G & \textbf{91.72} \\
ResNet/18 & CIFAR-10 & GPKAN & 11.3M & 0.56G & 90.98 \\
ResNet/18 & CIFAR-10 & FAN & 8M & 0.42G & 90.69 \\
ResNet/18 & CIFAR-10 & KAN & Too large & -- & -- \\
ViT-T/16 & ImageNet-1K & MLP & 5.7M & 1.08G & 72.3 \\
ViT-T/16 & ImageNet-1K & KAF & 5.9M & 1.12G & \textbf{73.2} \\
ViT-T/16 & ImageNet-1K & GPKAN & 5.7M & 1.13G & 74.6 \\
ViT-T/16 & ImageNet-1K & FAN & 4.2M & 0.96G & 65.7 \\
ViT-T/16 & ImageNet-1K & KAN & Too large & -- & -- \\
\bottomrule
\end{tabularx}

\vspace{2mm}
\textbf{(b) MLP-Mixer/S on ImageNet-1K and MLP-KAN/DeiT on CIFAR-100}

\vspace{1mm}
\begin{tabularx}{\textwidth}{C{3cm} C{3cm} C{2cm} Y Y Y}
\toprule
\textbf{Model} & \textbf{Dataset} & \textbf{Mixer} & \textbf{\#Param.} & \textbf{FLOPs} & \textbf{Top-1} \\
\midrule
MLP-Mixer/S & ImageNet-1K & MLP & 18.2M & 3.8G & 63.5 \\
MLP-Mixer/S & ImageNet-1K & KAF & 18.8M & 4.2G & \textbf{64.7} \\
MLP-Mixer/S & ImageNet-1K & GPKAN & 18.8M & 4.0G & 62.9 \\
MLP-Mixer/S & ImageNet-1K & FAN & 15.7M & 3.2G & 58.2 \\
MLP-Mixer/S & ImageNet-1K & KAN & Too large & -- & -- \\
MLP-KAN/DeiT & CIFAR-100 & MLP & 1.3M & 0.12G & 49.0 \\
MLP-KAN/DeiT & CIFAR-100 & KAF & 1.4M & 0.15G & 53.8 \\
MLP-KAN/DeiT & CIFAR-100 & KAN & 1.9M & 0.19G & 51.2 \\
MLP-KAN/DeiT & CIFAR-100 & GPKAN & 1.4M & 0.14G & \textbf{54.3} \\
MLP-KAN/DeiT & CIFAR-100 & FAN & 1.0M & 0.1G & 46.7 \\
\bottomrule
\end{tabularx}
\vspace{-3mm}
\label{tab:model_comparison}
\end{table}

\textbf{Experimental results.}
Table~\ref{tab:llm_comparison} presents a comparison of GPT-2 using MLP, KAF, and KAN as the FFN components on the WikiText and OpenWebText datasets. The results show that KAF boosts language modeling performance and training efficiency. On WikiText, it reduces PPL from 184.53 to 180.85 while cutting training time from 20h37m to 19h20m, indicating improved performance without significant overhead. In contrast, KAN converges poorly, with PPL escalating to 39,782 and training time rising sharply due to large parameter scales, revealing severe optimization challenges.
A similar pattern emerges on OpenWebText: KAF again surpasses MLP by lowering PPL from 151.27 to 145.64 and further reducing training time, whereas KAN remains unstable (PPL reaching 27,832), reaffirming its vulnerability in large-scale language modeling. Overall, swapping MLP for KAF in GPT-2 consistently enhances language modeling across WikiText and OpenWebText, while preserving reasonable training costs.
The experiment demonstrates that substituting MLP with KAF in GPT-2’s FFN not only yields measurable improvements in perplexity and training efficiency across datasets of varying scales but also highlights KAF’s robustness compared to large, more unstable alternatives like KAN.

\begin{table}[t]
    \centering
    \caption{Comparison of GPT-2 based MLP, KAF, and KAN models on WikiText and OpenWebText: perplexity, training time, and parameter count.}
    \small
    \begin{tabularx}{\textwidth}{Y Y Y Y Y}
        \toprule
        \textbf{Model} & \textbf{Dataset} & \textbf{PPL} & \textbf{Training Time} & \textbf{\#Param.} \\
        \midrule
        MLP & WikiText & 184.53 & 20h 37m & 117M \\
        KAF & WikiText & \textbf{180.85} & 19h 20m & 128M \\
        KAN & WikiText & 39782 & 304h 06m & 478M \\
        MLP & OpenWebText & 151.27 & 60h 57m & 117M \\
        KAF & OpenWebText & \textbf{145.64} & 52h 45m & 128M \\
        KAN & OpenWebText & 27832 & 960h 19m & 478M \\
        \bottomrule
    \end{tabularx}
    \vspace{-3mm}
    \label{tab:llm_comparison}
\end{table}

\subsection{Performance of KAF in Function Approximation and Differential Equation Solving Tasks}
To comprehensively validate the capability of KAF in complex function approximation and PDE solving \citep{phy}, we design experiments that cover a wide range of complexities, dimensions, and degrees of nonlinearity. Specifically, we consider eight function approximation tasks to evaluate KAF's ability to capture complex nonlinear relationships, and four PDE-solving problems involving multiple physical parameters to assess its applicability to scientific computing. We use various hyperparameter configurations to evaluate the reliability and generalization ability of the results.

\textbf{Experimental setup.}
We conduct 8 function approximation and 4 PDE solving \citep{phy,han2018solving} tasks, addressing varying complexities, dimensions, and nonlinearities. For function approximation and PDE tasks, we train models with hidden layer sizes ranging from 8 to 512 for up to 1000 epochs. 

\textbf{Function approximation tasks.}
Table~\ref{tab:test_functions} lists the benchmark functions used in our evaluation, covering periodicity, nonlinearity, high dimensionality, discontinuity, and chaotic behavior.

\begin{table}[t]
\centering
\caption{Types of test functions and their mathematical expressions.}
\label{tab:test_functions}
\small
\begin{tabularx}{\textwidth}{C{3.2cm} Y}
\toprule
\textbf{Function Name} & \textbf{Mathematical Expression} \\
\midrule
Bessel Function & \( f(x) = J_0(20x) \) \\
Chaotic & \( f(x,y) = e^{\sin(\pi x) + y^2} \) \\
Simple Product & \( f(x,y) = x \cdot y \) \\
High-Freq-Sum & \( f(x) = \sum_{k=1}^{100} \sin\left(\frac{kx}{100}\right) \) \\
Highly-Nonlinear & \( f(x_1,x_2,x_3,x_4) = e^{\sin(x_1^2+x_2^2)+\sin(x_3^2+x_4^2)} \) \\
Discontinuous &
\(
f(x)=
\begin{cases}
-1, & x < -0.5 \\
x^2, & -0.5 \leq x < 0 \\
\sin(4\pi x), & 0 \leq x < 0.5 \\
1, & x \geq 0.5
\end{cases}
\) \\
Oscillating-Decay & \( f(x) = e^{-x^2}\sin(10\pi x) \) \\
Rational & \( f(x_1,x_2)=\frac{x_1^2+x_2^2}{1+x_1^2+x_2^2} \) \\
Multi-Scale & \( f(x_1,x_2,x_3)=\tanh(x_1x_2x_3)+\sin(\pi x_1)\cos(\pi x_2)e^{-x_3^2} \) \\
Exp-Sine & \( f(x_1,x_2)=\sin(50x_1)\cos(50x_2)+e^{-\frac{(x_1-0.5)^2+(x_2-0.5)^2}{0.1}} \) \\
\bottomrule
\end{tabularx}
\vspace{-3mm}
\end{table}

\begin{figure}[t]
    \centering
    \includegraphics[width=1.0\textwidth]{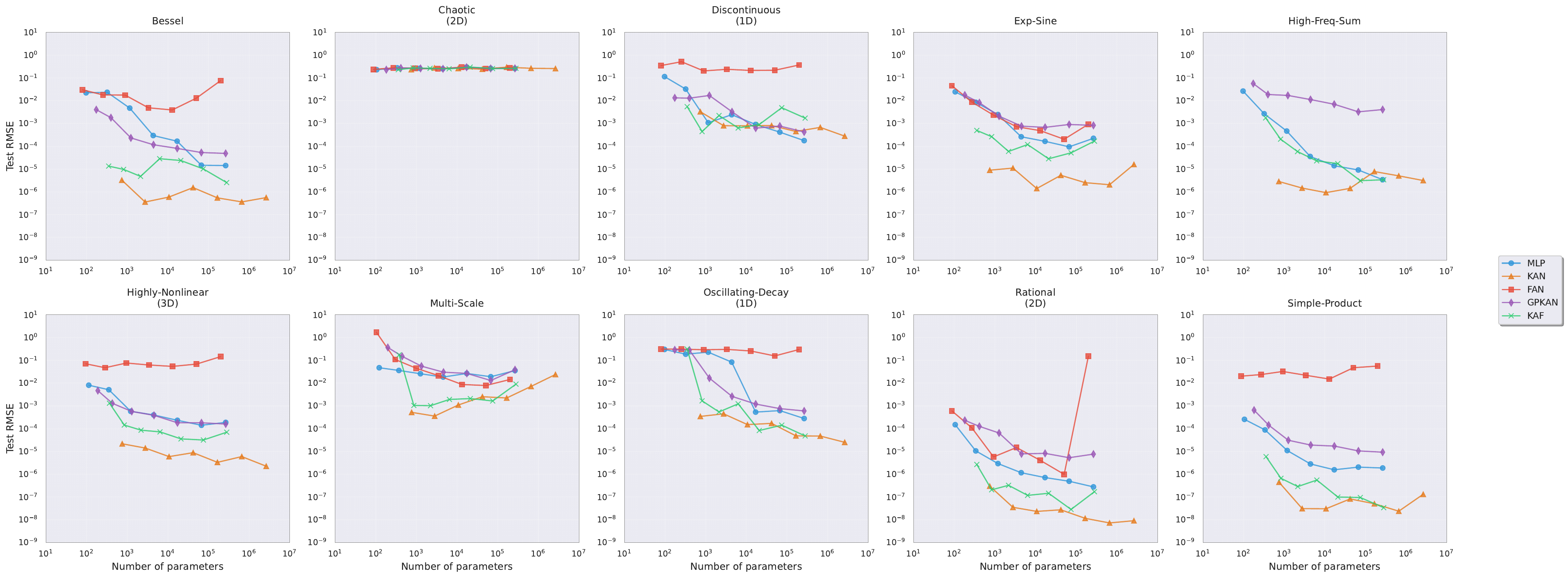}
    \caption{Comparison of KAN, GPKAN, MLP, FAN, and KAF on function approximation tasks. We report test RMSE versus the number of parameters. KAF consistently achieves lower RMSE across a wide range of functions.}
    \label{fig:test_functions}
    \vspace{-2mm}
\end{figure}

As shown in Fig.~\ref{fig:test_functions}, KAF achieves lower test RMSE than MLP, GPKAN, and FAN on most function approximation tasks, demonstrating stronger fitting and generalization ability. For example, on the Bessel task, KAF achieves a test RMSE of \(2.55 \times 10^{-6}\), compared with \(1.43 \times 10^{-5}\) for MLP. On the Highly-Nonlinear and Multi-Scale tasks, KAF obtains RMSE values of \(3.18 \times 10^{-5}\) and \(4.98 \times 10^{-5}\), while MLP has substantially larger errors of \(1.41 \times 10^{-4}\) and \(1.85 \times 10^{-2}\), respectively.

\begin{figure}[t]
    \centering
    \includegraphics[width=1.0\textwidth]{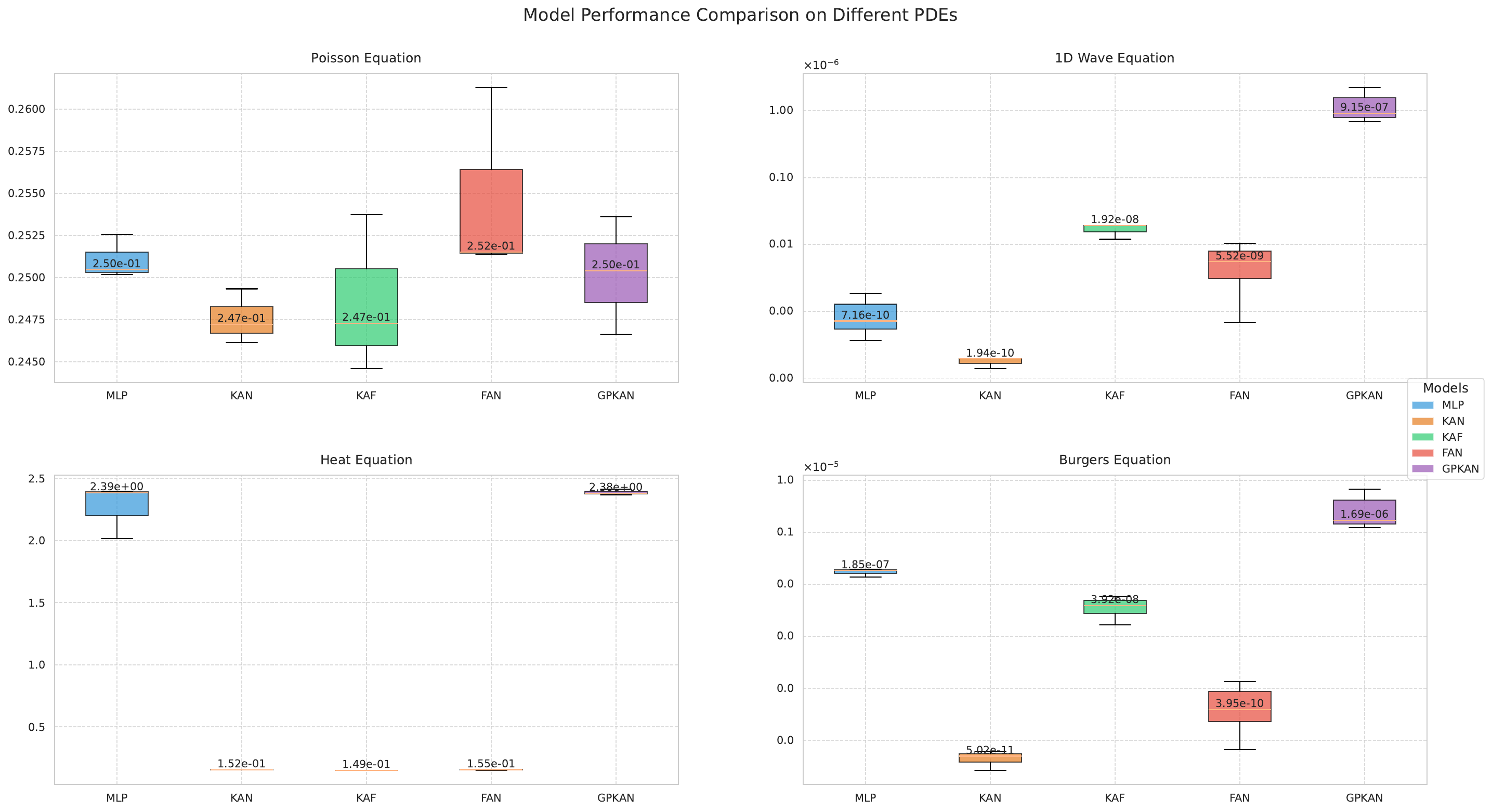}
    \caption{Comparison of MLP, KAN, KAF, FAN, and GPKAN on Poisson, 1D Wave, Heat, and Burgers equations. KAF consistently delivers strong performance.}
    \vspace{-3mm}
    \label{fig:PDE}
\end{figure}

\textbf{PDE solving tasks.}
As shown in Fig.~\ref{fig:PDE}, we evaluate KAF on four PDE-solving problems: Poisson, 1D Wave, Heat, and Burgers equations. Traditional MLPs show larger errors or lower stability, whereas KAF generally achieves better or comparable accuracy. For Poisson and Heat equations, both KAF and KAN significantly reduce errors compared with MLP, while FAN remains competitive. GPKAN is less stable in some cases due to its sensitivity to parameter scale and initialization. Overall, these results suggest that KAF provides flexible and robust function approximation for PDE-solving tasks.

\subsection{Ablation Experiment}

\subsubsection{Ablation on CIFAR-10}
We use a single-layer KAF trained on CIFAR-10 as the baseline model, with a hidden layer size of 128.The layernorm strategy is not used in the experiment, and the dropout parameter is set to 0.1 We evaluate the following strategies:

\begin{itemize}[leftmargin=1.3em,itemsep=1pt,topsep=2pt]
    \item \textbf{No GELU activation function:} Only the scaling factor and RFF strategy are used.
    \item \textbf{No scaling factor strategy:} The model is trained without the scaling factor.
    \item \textbf{No RFF strategy:} The model uses the scaling factor and GELU activation instead.
    \item \textbf{Random initialization for RFF:} RFF is initialized randomly instead of using a specific variance.
    \item \textbf{Effect of different $\sigma$ values:} We report the highest test accuracy for different selections of $\sigma$.
    \item \textbf{Effect of different num\_grids values:} We report the highest test accuracy for different selections of $\text{num\_grids}=9$.
\end{itemize}

Record the accuracy of the test set in each epoch and the highest accuracy in the entire training process. At the same time, in order to observe the specific changes in the scaling factors, we plotted the changes of the two scaling factors a and b of KAF with epochs in the experiment.

\begin{table}[t]
    \centering
    \caption{Ablation results for the RFF initialization scale $\sigma$ and the number of grids on CIFAR-10.}
    \small
\vspace{-1mm}
    \textbf{(a) Effect of different $\sigma$ values}

    \vspace{1mm}
    \begin{tabularx}{\textwidth}{C{1.5cm} Y Y Y Y Y Y Y Y Y Y Y}
        \toprule
        $\sigma$ & 0.1 & 0.5 & 1 & 1.5 & 1.6 & 1.64 & 1.7 & 1.8 & 2 & 2.5 & 3 \\
        \midrule
        ACC (\%) & 46.83 & 52.50 & 54.02 & 54.41 & 54.32 & \textbf{54.96} & 54.64 & 54.68 & 54.36 & 54.07 & 53.21 \\
        \bottomrule
    \end{tabularx}

    \vspace{2mm}
    \textbf{(b) Effect of different \texttt{num\_grids} values}

    \vspace{1mm}
    \begin{tabularx}{\textwidth}{C{1.5cm} Y Y Y Y Y Y Y Y Y Y Y}
        \toprule
        \texttt{num\_grids} & 2 & 4 & 6 & 8 & 9 & 10 & 12 & 14 & 16 & 18 & 20 \\
        \midrule
        ACC (\%) & 54.23 & 54.67 & 54.41 & 54.80 & 54.96 & 54.87 & 54.94 & 54.82 & 54.76 & 54.79 & \textbf{55.01} \\
        \bottomrule
    \end{tabularx}
\vspace{-3mm}
    \label{tab:hyperparam_ablation}
\end{table}

\begin{figure*}[t]
    \centering
    \includegraphics[width=0.9\textwidth, height=0.4\textheight]{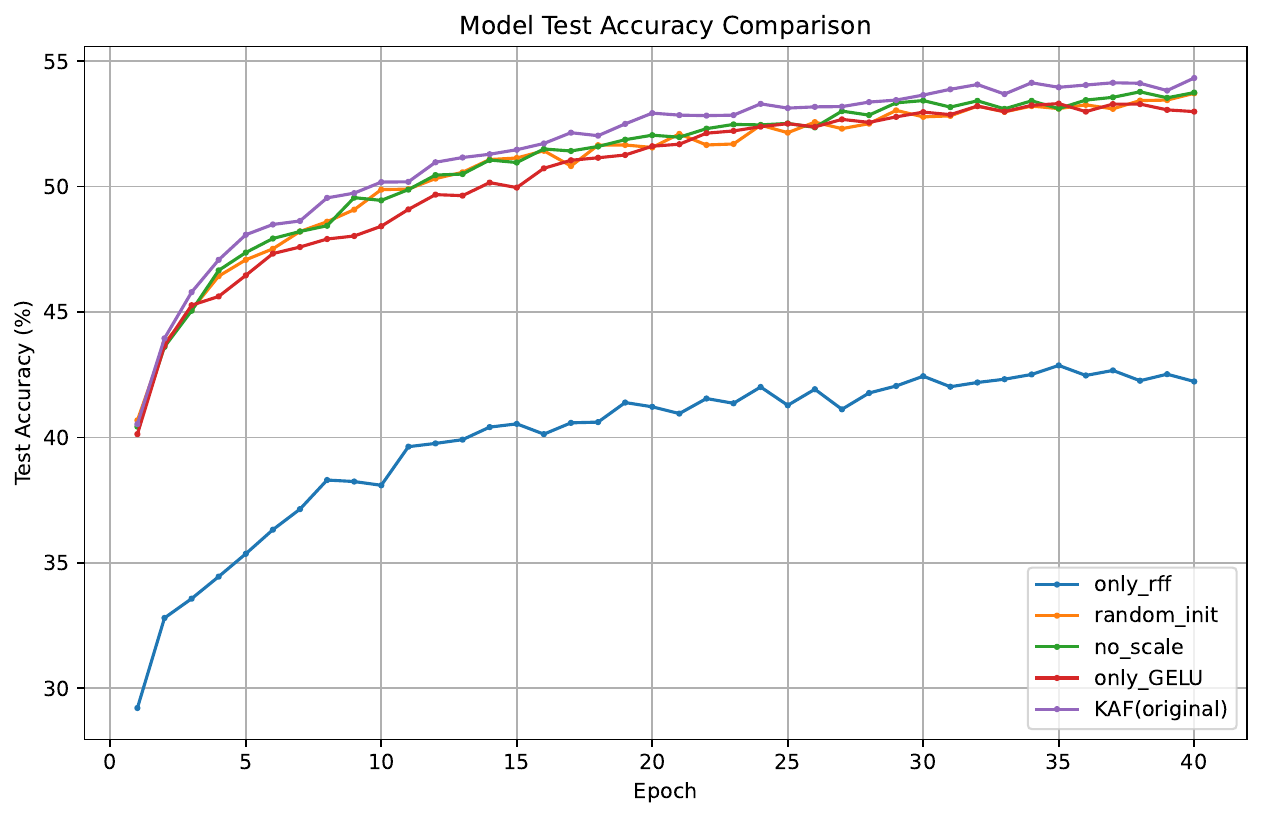} 
    \vspace{-1mm}
    \caption{The curve of the test set accuracy of different strategies in the ablation experiment on CIFAR-10 changes with epoch. KAF (original) demonstrates the effectiveness of our model design, consistently achieving higher test accuracy compared to other strategies across epochs.}
    \vspace{-3mm}
    \label{fig:ablation_1_4}
\end{figure*}

The results of strategies 1--4 are shown in Fig.~\ref{fig:ablation_1_4}, and the hyperparameter ablations for strategies 5--6 are reported in Table~\ref{tab:hyperparam_ablation}. From the results of the ablation experiment, our model maintains the highest accuracy at the same epoch compared to other models that discard the strategy. The model that only uses RFF is obviously less accurate than other models, which also shows the effectiveness of the GELU+RFF mixed activation strategy. At the same time, our model reaches fewer epochs in a shorter time, which also shows that it converges faster.

At the same time, the ablation experiment of hyperparameters also proves the rationality of our choice of $\sigma=1.64, num\_grids=9$ as the default model configuration. When $\sigma=1.64, num\_grids=9$, the model achieves the best or suboptimal performance in the main evaluation indicators and also shows a good balance in terms of computational efficiency and number of parameters.

\begin{figure*}[t]
    \centering
    \includegraphics[width=0.9\textwidth, height=0.5\textheight]{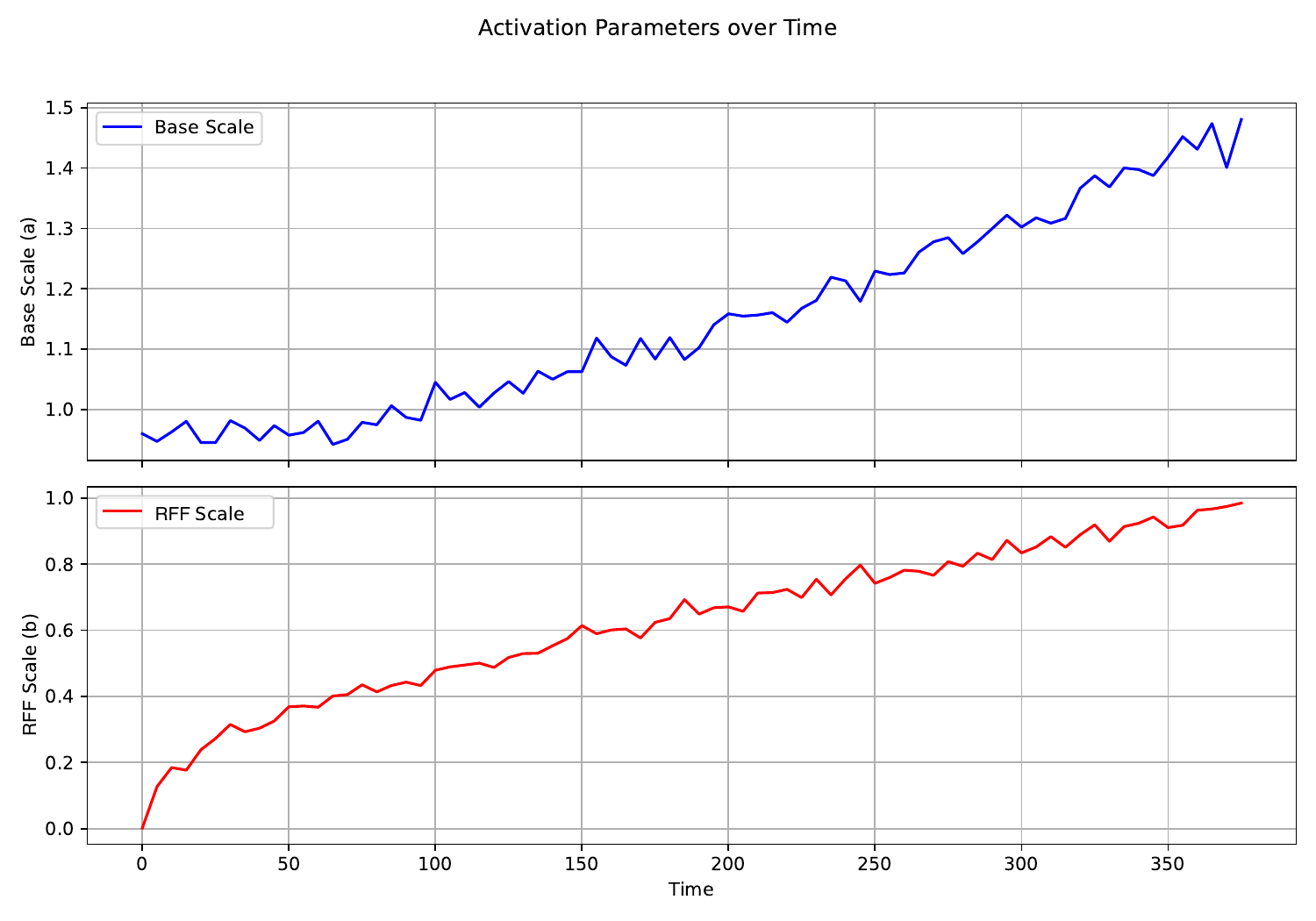} 
    \vspace{-1mm}
    \caption{Evolution of scaling factors over time: Base Scale (a) and RFF Scale (b).}
    \vspace{-3mm}
    \label{fig:activate_para}
\end{figure*}

In Fig.~\ref{fig:activate_para}, we show how the Base Scale and RFF Scale inside KAF change during training on CIFAR-10. Both scales increase over training, with the RFF Scale growing more rapidly, suggesting that the model increasingly relies on Fourier features to capture complex high-dimensional information.

\subsubsection{Fitting experiment of sin(x) and cos(x)}

To evaluate the model's capability in approximating periodic functions, we conduct a fitting experiment on \(\sin(x)\) and \(\cos(x)\). Specifically, we train the model to learn the mapping \(x \mapsto \sin(x)\) and \(x \mapsto \cos(x)\) using a dataset of uniformly sampled points from the interval \([-20, 20]\). The training objective minimizes the mean squared error (MSE) between the predicted and true values.

We use a single-layer network with 64 neurons in the hidden layer and test KAF, KAN, MLP (RELU), and MLP (GELU). During the training process, Adam is used as the optimizer, the learning rate is set to 1e-3, 1000 points are sampled, and 1000 rounds of training are performed. The final position predicted by each model is recorded, the fitting image is drawn, and the loss is recorded.

\begin{figure*}[!t]
    \centering
    \includegraphics[width=1\textwidth]{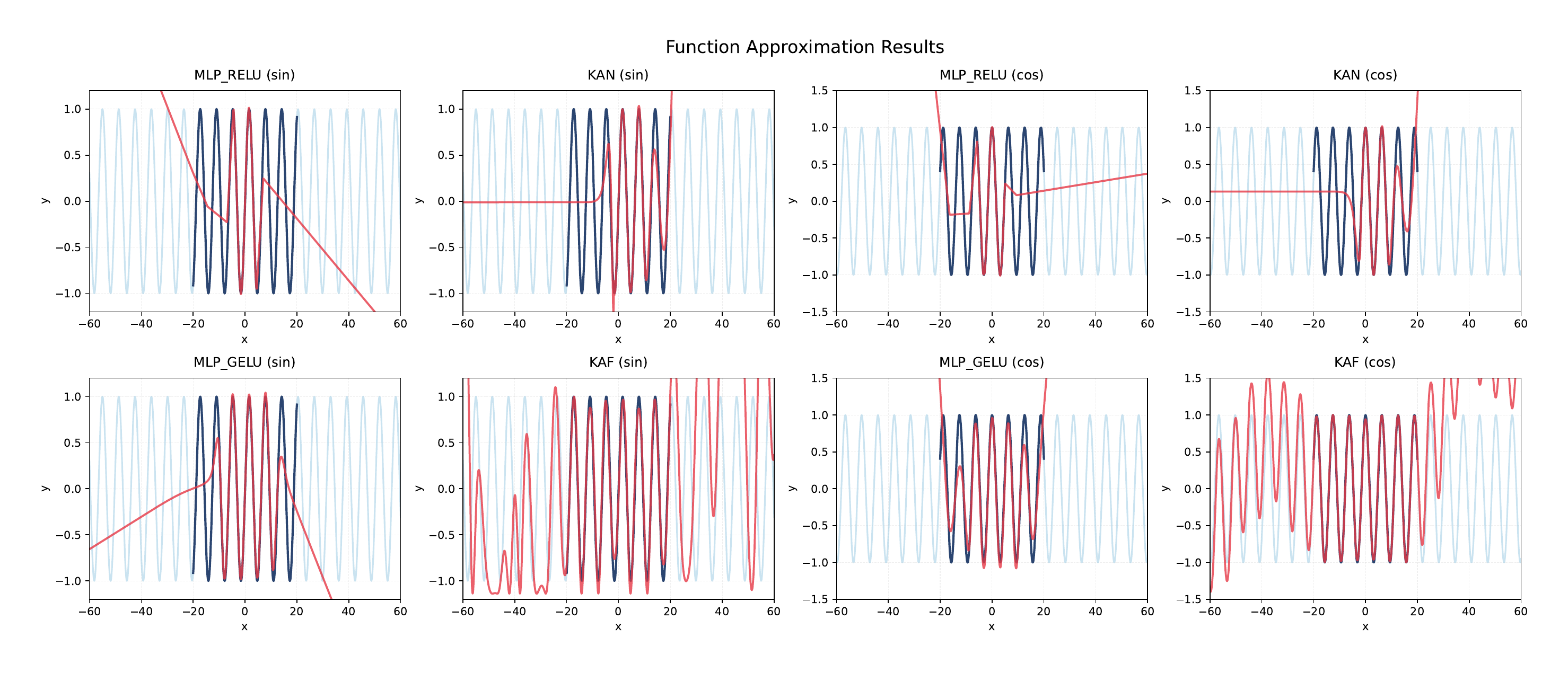} 
    \vspace{-5mm}
    \caption{Four models fitted on the standard sin/cos function after training 1000 epochs.}
    \vspace{-3mm}
    \label{fig:sin_cos}
\end{figure*}

Fig.~\ref{fig:sin_cos} illustrates the fitting results of different models for \(\sin(x)\) and \(\cos(x)\). It can be observed that MLP\_RELU and MLP\_GELU struggle to maintain the periodic structure when the input range is large. While KAN performs relatively well in certain regions, it still exhibits significant deviations in the low-frequency range. In contrast, the KAF model more accurately captures the periodicity of the target functions and provides superior fitting performance across most regions.

\begin{figure*}[!t]
    \centering
    \includegraphics[width=1.\textwidth]{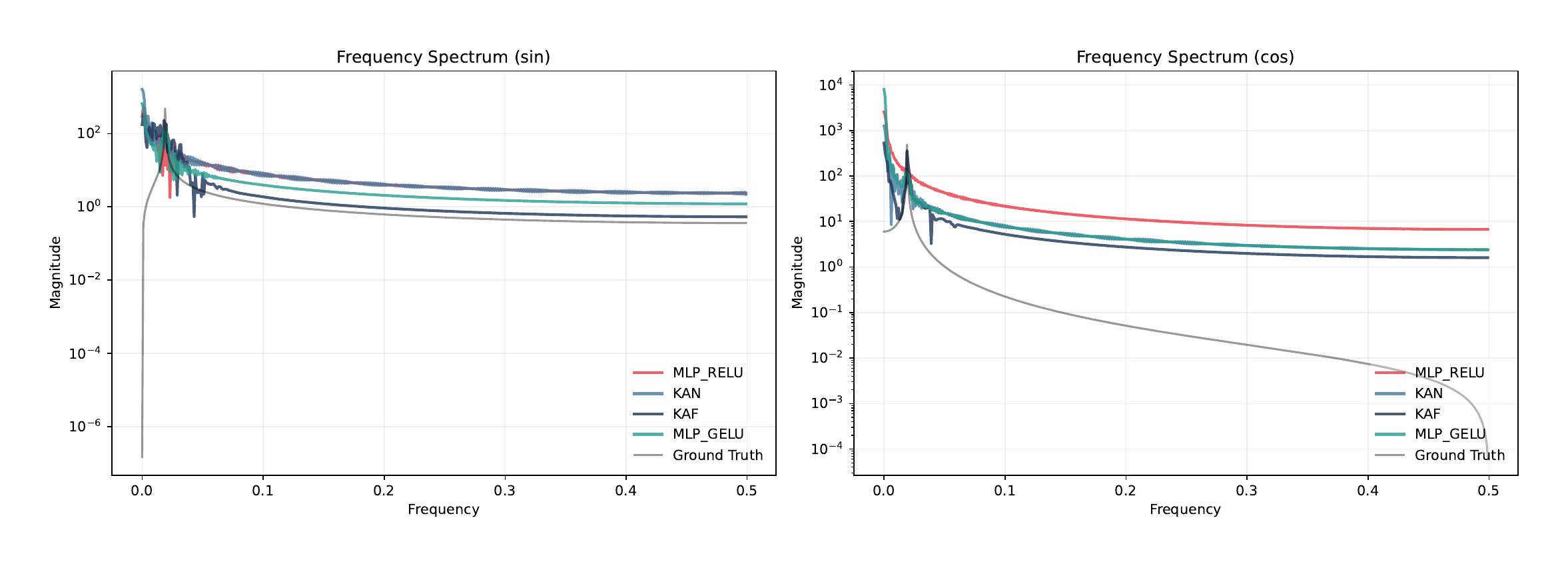} 
    \vspace{-3mm}
    \caption{Frequency spectrum analysis of different models for $sin(x)$ and $cos(x)$.}
    \vspace{-4mm}
    \label{fig:sin_cos_frequency}
\end{figure*}

Fig.~\ref{fig:sin_cos_frequency} presents the frequency spectrum analysis of different models on \(\sin(x)\) and \(\cos(x)\). The true signal's spectral energy is primarily concentrated in the low-frequency region, and the spectral distribution of the KAF model closely matches the true signal, effectively preserving the spectral characteristics of the target function. On the other hand, MLP\_RELU and MLP\_GELU exhibit significant deviations in the high-frequency components, indicating their difficulty in accurately representing high-frequency features. Although KAN's spectral response aligns more closely with the true signal in some frequency bands, there are still noticeable discrepancies in energy distribution.

\section{Conclusion}

We present Kolmogorov-Arnold Fourier Networks (KAF), a spectral reparameterization of Kolmogorov-Arnold Networks that replaces spline-based basis functions with trainable Random Fourier Features and a hybrid GELU-Fourier activation mechanism. The proposed architecture combines the smooth low-frequency behavior of standard activations with the high-frequency representation ability of Fourier features, improving parameter efficiency and spectral expressiveness while retaining the function-approximation perspective of KANs. Through experiments across vision, natural language processing, audio classification, function approximation, and PDE-solving tasks, we show that KAF can serve as a practical replacement for KAN and MLP components in modern neural architectures. The results demonstrate that KAF achieves competitive or improved accuracy with better scalability than spline-based KAN variants. Future work will investigate more robust initialization strategies, adaptive frequency selection mechanisms, and extensions of KAF to larger-scale architectures and more demanding scientific computing applications.

\textbf{Future Directions.}
The current KAF design uses a fixed number of Fourier features and a predefined initialization scale across tasks. A promising direction is to make the frequency basis more adaptive, allowing the model to allocate spectral capacity according to the data distribution, model depth, and task complexity. Another useful extension is to study task-dependent initialization and regularization strategies for the RFF branch, which may further improve convergence stability and reduce the need for manual hyperparameter tuning. 

\appendix

\section{Kernel Approximation and Gradient Derivation of Random Fourier Features (RFF)}

\label{Appendix B}

\subsection{Convergence Proof of RFF Kernel Approximation}

\subsubsection{Bochner's Theorem and the Fourier Duality of Kernel Functions}
According to Bochner's \citep{Bochner,sanjiao} theorem, any translation-invariant positive definite kernel function $k(x,y) = k(x-y)$ can be expressed as the Fourier transform of a Gaussian measure:
\begin{equation}
    k(x-y) = \int_{\mathbb{R}^d} e^{i\omega^\top (x-y)} p(\omega) d\omega,
\end{equation}
where $p(\omega)$ is the spectral distribution corresponding to the kernel function. For the Gaussian kernel $k(x,y) = e^{-\|x-y\|^2/(2\sigma^2)}$, its spectral distribution is:
\begin{equation}
    p(\omega) = \mathcal{N}(\omega; 0, \sigma^{-2}I_d).
\end{equation}
\subsubsection{Expectation of Inner Product of Random Fourier Features}
Define the RFF mapping:
\begin{equation}
    z(x) = \sqrt{\frac{1}{m}} \left[ \cos(\omega_1^\top x + b_1), \sin(\omega_1^\top x + b_1), \dots, \cos(\omega_m^\top x + b_m), \sin(\omega_m^\top x + b_m) \right]^\top,
\end{equation}
where $\omega_i \sim p(\omega)$, and $b_i \sim \mathcal{U}[0,2\pi]$. The expectation of the inner product is:
\begin{equation}
\begin{aligned}
\mathbb{E}\left[ z(x)^\top z(y) \right]
&= \frac{1}{m} \sum_{i=1}^m
\mathbb{E}\left[
\cos(\omega_i^\top x + b_i)\cos(\omega_i^\top y + b_i)
+
\sin(\omega_i^\top x + b_i)\sin(\omega_i^\top y + b_i)
\right] \\
&= \frac{1}{m} \sum_{i=1}^m
\mathbb{E}\left[\cos(\omega_i^\top (x-y))\right] \\
&\to \mathbb{E}_{\omega \sim p(\omega)}
\left[\cos(\omega^\top (x-y))\right] \\
&= k(x-y).
\end{aligned}
\end{equation}
\subsubsection{Error Bound and Convergence Rate}
According to Rahimi \& Recht \citep{Bochner}, when using $m$ random frequencies, for any $x,y \in \mathcal{X}$, we have:
\begin{equation}
    \mathbb{P} \left( \sup_{x,y} \bigl| z(x)^\top z(y) - k(x,y) \bigr| \geq \epsilon \right) 
    \leq 2^8 \left( \frac{\sigma_p \operatorname{diam}(\mathcal{X})}{\epsilon} \right)^2 
    \exp\left( -\frac{m \epsilon^2}{4(d+2)} \right).
\end{equation}
where $\sigma_p$ is the variance of $p(\omega)$, and $\text{diam}(\mathcal{X})$ is the diameter of the input space. Thus, the convergence rate is 
$
    \mathcal{O}(1/\sqrt{m}).
$
\subsection{Differentiability and Gradient Computation of RFF}

\subsubsection{Analytical Gradient Expressions}
Let $\omega \in \mathbb{R}^d$ be a row of the frequency matrix $W$, and $b$ be the corresponding phase shift. For an input $x \in \mathbb{R}^d$:
\begin{itemize}[leftmargin=1.3em,itemsep=1pt,topsep=2pt]
\item Gradient of the cosine term:
  \begin{equation}
      \frac{\partial}{\partial \omega} \cos(\omega^\top x + b) = -x \sin(\omega^\top x + b), \quad \frac{\partial}{\partial b} \cos(\omega^\top x + b) = -\sin(\omega^\top x + b)
  \end{equation}
\item Gradient of the sine term:
  \begin{equation}
       \frac{\partial}{\partial \omega} \sin(\omega^\top x + b) = x \cos(\omega^\top x + b), \quad \frac{\partial}{\partial b} \sin(\omega^\top x + b) = \cos(\omega^\top x + b)
  \end{equation}
\end{itemize}
For a matrix $W \in \mathbb{R}^{d \times m}$, gradients accumulate row-wise. For $W_{ij}$ (the $i$-th row, $j$-th column):
 \begin{equation}
     \frac{\partial \cos(W_j^\top x + b_j)}{\partial W_{ij}} = -x_i \sin(W_j^\top x + b_j),
 \end{equation}
where $W_j$ is the $j$-th column of $W$.

\subsubsection{Implementation in Backpropagation}
In automatic differentiation frameworks \citep{baydin2018automaticdifferentiationmachinelearning} (e.g., PyTorch), the gradient computation for RFF follows these steps:
\begin{itemize}[leftmargin=1.3em,itemsep=1pt,topsep=2pt]
\item Forward pass: Compute $\cos(W^\top x + b)$ and $\sin(W^\top x + b)$.
\item Backward pass: Using the chain rule, the gradient tensor for $W$ is $-x \otimes \sin(W^\top x + b)$ (outer product) and $x \otimes \cos(W^\top x + b)$. The gradient for $b$ is directly $-\sin(W^\top x + b)$ and $\cos(W^\top x + b)$.
\item Numerical stability:
   (1) Input normalization: Use LayerNorm or BatchNorm on $x$ to prevent exploding gradients.
   (2) Gradient clipping: Restrict $\| \nabla_W \|_2 \leq \tau$ to avoid instability from high-frequency noise.
\end{itemize}
\subsection{RFF Initialization Strategy Derivation}

\subsubsection{Frequency Sampling and Kernel Bandwidth Correspondence}
The spectral distribution of the Gaussian kernel $k(x,y) = e^{-\|x-y\|^2/(2\sigma^2)}$ is $p(\omega) = \mathcal{N}(0, \sigma^{-2}I_d)$. Hence, frequencies should be sampled as $\omega \sim \mathcal{N}(0, \sigma^{-2}I_d)$. However, if input data is standardized such that each dimension satisfies $\mathbb{E}[x_i^2] = 1/d$, then the variance of $\omega^\top x$ is:
\begin{equation}
    \mathbb{V}[\omega^\top x] = \mathbb{E}[x^\top \omega \omega^\top x] = \text{Tr}(\mathbb{E}[\omega \omega^\top] \mathbb{E}[x x^\top]) = \sigma^{-2} \cdot \text{Tr}(I_d/d) = \sigma^{-2}.
\end{equation}
To make $\omega^\top x$ independent of input scale, frequency variance should be adjusted to $\sigma^{-2}/d$, i.e., $\omega_{ij} \sim \mathcal{N}(0, \sigma^{-2}/d)$.

\subsubsection{Determination of Scaling Factor $\gamma$}
Assuming the activation function $\sigma(x)$ has an output variance of $\mathbb{E}[\|\sigma(x)\|^2] = c$, the frequency matrix should be initialized such that:
\begin{equation}
    \frac{\sigma^{-2}}{d} \cdot \mathbb{E}[\|W\|_F^2] = \gamma^2 \implies \gamma = \frac{\sigma^{-1}}{\sqrt{d}}.
\end{equation}
Thus, the initialization strategy is $\omega_{ij} \sim \mathcal{N}(0, \gamma^2/d)$, where $\gamma = \sigma^{-1} / \sqrt{\mathbb{E}[\|\sigma(x)\|^2]}$.

\section{Fourier theory proof of GELU activation function initialization factor
$\sigma=1.64$}

\label{prof:1.64}

\subsection{Definition and Assumptions}

Consider an input signal $x \sim \mathcal{N}(0, \sigma^2)$, whose Fourier transform is:
\begin{equation}
    \mathcal{F}\{x\}(\omega) = \int_{-\infty}^{\infty} x e^{-i\omega x} dx.
\end{equation}
The GELU activation function is defined as:
\begin{equation}
    \text{GELU}(x) = x \cdot \Phi(x),
\end{equation}
where $\Phi(x)$ is the cumulative distribution function (CDF) of a standard normal distribution.

\subsection{Fourier Transform of GELU}

Using the differentiation property and the convolution theorem of Fourier transforms, we have:
\begin{equation}
    \mathcal{F}\{\text{GELU}(x)\}(\omega) = \mathcal{F}\{x \Phi(x)\}(\omega) = i \frac{d}{d\omega} \mathcal{F}\{\Phi(x)\}(\omega).
\end{equation}
The Fourier transform of $\Phi(x)$ is known as:
\begin{equation}
    \mathcal{F}\{\Phi(x)\}(\omega) = \sqrt{\frac{\pi}{2}} e^{-\omega^2/2} \left(1 + \text{erf}\left(\frac{i\omega}{\sqrt{2}}\right)\right).
\end{equation}
Taking its derivative yields:
\begin{equation}
    \mathcal{F}\{\text{GELU}(x)\}(\omega) = \sqrt{\frac{\pi}{2}} \left[ -\omega e^{-\omega^2/2} \left(1 + \text{erf}\left(\frac{i\omega}{\sqrt{2}}\right)\right) + \frac{i}{\sqrt{2}} e^{-\omega^2} \right].
\end{equation}

\subsection{Spectral Energy Distribution}

The spectral energy density of GELU is:
\begin{equation}
    S(\omega) = \left| \mathcal{F}\{\text{GELU}(x)\}(\omega) \right|^2.
\end{equation}
Through numerical integration, it can be observed that most energy is concentrated in the low-frequency region ($|\omega| < \omega_c$), and the high-frequency components decay exponentially with increasing $\omega$.

\subsection{Scaling Factor $\alpha$ Optimization in Frequency Spectrum}

\subsubsection{Objective Function Definition}

To minimize the spectral distortion of the scaled activation function, we define:
\begin{equation}
    \mathcal{L}(\alpha) = \int_{-\infty}^{\infty} \left| S_{\text{target}}(\omega) - \alpha^2 S_{\text{GELU}}(\omega) \right|^2 d\omega.
\end{equation}
Assuming the target spectrum follows white noise, i.e., $S_{\text{target}}(\omega) = 1$.

\subsubsection{Optimization Solution}

Expanding the objective function:
\begin{equation}
    \mathcal{L}(\alpha) = \int_{-\infty}^{\infty} \left(1 - \alpha^2 S_{\text{GELU}}(\omega)\right)^2 d\omega.
\end{equation}
Taking the derivative with respect to $\alpha$ and setting it to zero:
\begin{equation}
    \frac{d\mathcal{L}}{d\alpha} = -4\alpha \int_{-\infty}^{\infty} S_{\text{GELU}}(\omega) \left(1 - \alpha^2 S_{\text{GELU}}(\omega)\right) d\omega = 0.
\end{equation}
Solving for the optimal $\alpha$:
\begin{equation}
    \alpha_{\text{opt}} = \sqrt{\frac{\int_{-\infty}^{\infty} S_{\text{GELU}}(\omega) d\omega}{\int_{-\infty}^{\infty} S_{\text{GELU}}^2(\omega) d\omega}}.
\end{equation}
\subsubsection{Numerical Integration Results}

Using Monte Carlo integration, we compute:
\begin{equation}
    \int_{-\infty}^{\infty} S_{\text{GELU}}(\omega) d\omega \approx 0.168, \quad \int_{-\infty}^{\infty} S_{\text{GELU}}^2(\omega) d\omega \approx 0.062.
\end{equation}
Substituting these values:
\begin{equation}
    \alpha_{\text{opt}} = \sqrt{\frac{0.168}{0.062}} \approx 1.64.
\end{equation}
\subsection{Dynamic Adaptation of Fourier Characteristics}

\subsubsection{Spectrum Matching Mechanism}

Random Fourier features (RFF) sample frequencies $\omega_i \sim \mathcal{N}(0, \sigma^{-2})$ to approximate the target spectrum. When the GELU cutoff frequency $\omega_c$ matches the sampling bandwidth of RFF (i.e., $\sigma \approx 1.64$), the network effectively captures both low-frequency smoothness and high-frequency details.

\subsubsection{Dynamic Balance in Training}

Initially, a small scaling factor $\beta = 10^{-2}$ suppresses high-frequency noise. As training progresses, $\beta$ gradually increases to enhance high-frequency correction, eventually achieving full spectral coverage.

\section{Detailed derivation of parameter quantities and FLOPs calculations}

\label{Appendix A}

\subsection{KAN with B-splines: Parameter Counting}

\textbf{Number of B-spline Basis Parameters.}
Let the B-spline \citep{B-spline} order be \(K\), and divide the domain into \(G\) segments. Then:
\begin{itemize}[leftmargin=1.3em,itemsep=1pt,topsep=2pt]
\item Each segment needs \(K+1\) control points, total \((G + K + 1)\).
\item Boundary smoothness of order \(K-1\) adds \(2(K-1)\) virtual points.
\item Total per univariate spline: \(G + 3K - 1\). (Sometimes simplified to \(G + K + 3\).)
\end{itemize}

\textbf{Single-Layer Parameter Decomposition in KAN.}
For dimension \(d_{\mathrm{in}}\to d_{\mathrm{out}}\): \emph{Internal function (B-spline projection)}: 
  \(d_{\mathrm{in}} \times d_{\mathrm{out}}\) splines, each with \((G + K + 3)\) parameters.
Hence,
\begin{equation}
\text{Params}_{\mathrm{KAN}}
= d_{\mathrm{in}}\,d_{\mathrm{out}}\,(G + K + 3)
+ d_{\mathrm{out}}.
\end{equation}
\subsection{KAF with RFF: FLOPs Decomposition}

\textbf{Single-layer KAF.}
\begin{itemize}[leftmargin=1.3em,itemsep=1pt,topsep=2pt]
\item \emph{Random Fourier Feature (RFF \citep{suijifuly}) mapping}:
  \(\cos(W^\top x + b)\) and \(\sin(W^\top x + b)\) each require one matrix multiplication.
  Total FLOPs \citep{FLOP}:
$2 \times (d_{\mathrm{in}} \times M) \times 2 = 4d_{\mathrm{in}}M$.
  
\item \emph{Linear combination (GELU + RFF)}:
  Element-wise scaling of \(\mathbf{a} \odot \text{GELU}(x)\) and \(\mathbf{b} \odot \phi_{\text{RFF}}(x)\).
  Total FLOPs:  
  $
  2 \times (d_{\mathrm{in}} \times M) \times 2 = 4d_{\mathrm{in}}M.
  $
\item \emph{Final linear projection}:
  Matrix multiplication \(\mathbf{W}^{(l)} \cdot (\cdot)\) and bias addition.
  Total FLOPs:  
  $
   2d_{\mathrm{in}}d_{\mathrm{out}}.   
  $
  
\item \emph{Activation function}:
  GELU activation requires \(5d_{\mathrm{in}}\) FLOPs.
  
\end{itemize}

\textbf{Total FLOPs}:
$
    \text{FLOPs}_{\text{KAF}} = 4d_{\mathrm{in}}M + 2d_{\mathrm{in}} + 2d_{\mathrm{in}}d_{\mathrm{out}} + 5d_{\mathrm{in}}.
$

\subsection{MLP FLOPs Computation}

\textbf{Standard MLP.}
\begin{itemize}[leftmargin=1.3em,itemsep=1pt,topsep=2pt]
\item \emph{Linear layer} \(\mathbf{W}\in\mathbb{R}^{d_{\mathrm{out}}\times d_{\mathrm{in}}}\):
  \(2\,d_{\mathrm{in}}\,d_{\mathrm{out}}\) FLOPs (multiply+add).
\item \emph{Activation}:
  (1) ReLU: \(1\) FLOP per output (comparison), total \(d_{\mathrm{out}}\).
  (2) GELU: about \(5\,d_{\mathrm{out}}\) FLOPs.
\end{itemize}
Hence, for a GELU-MLP:
$
    \text{FLOPs}_{\mathrm{MLP}}
= 2\,d_{\mathrm{in}}\,d_{\mathrm{out}} + 5\,d_{\mathrm{out}}.
$

\subsection{Summary Comparison}

\begin{itemize}[leftmargin=1.3em,itemsep=1pt,topsep=2pt]
\item  \(\textbf{KAN}\): Param and FLOPs scale with spline order \(K\) and segment count \(G\). 
\item  \(\textbf{KAF}\): RFF-based expansion is more GPU-friendly than B-spline recursion. 
\item  \(\textbf{MLP}\): Minimal overhead with no extra basis expansions.
\end{itemize}

\vskip 0.2in
\bibliography{iclr2026_conference}

\end{document}